\documentclass{article}
\usepackage{spconf,amsmath,graphicx,hyperref}
\usepackage{todonotes}
\usepackage{tikz}
\usepackage{graphicx}
\usepackage{amsmath}
\usepackage{pgfplots}
\usepackage{booktabs}
\usepackage[shortcuts,acronym]{glossaries}
\usetikzlibrary{positioning,arrows.meta,shapes.geometric,fit,calc}

\newacronym{ml}{ML}{Machine Learning}
\newacronym{nn}{NN}{Neural Networks}
\newacronym{fmcw}{FMCW}{Frequency-Modulated Continuous Wave}
\newacronym{rbs}{RBS}{Rule-Based System}
\newacronym{iai}{IAI}{Interpretable AI}
\newacronym{shap}{SHAP}{Shapley Additive Explanations}
\newacronym{xai}{XAI}{explainable AI}
\newacronym{er}{ER}{experience replay}
\newacronym{tl}{TL}{Transfer learning}
\newacronym{gru}{GRU}{gated recurrent unit}
\newacronym{hgr}{HGR}{hand gesture recognition}
\newacronym{mmwave}{mmWave}{millimeter-wave}
\newacronym{cnn}{CNN}{convolutional neural network}
\newacronym{lstm}{LSTM}{long short-term memory}
\newacronym{svm}{SVM}{Suport vector machines}
\newacronym{knn}{kNN}{k-nearest neighbors}
\newacronym{hmm}{HMM}{Hidden markov models}
\newacronym{pca}{PCA}{principle component analysis}
\newacronym{resnet}{ResNet}{Residual networks}
\newacronym{rmd}{RMD}{relative Mahalanobis distance}
\newacronym{gan}{GAN}{generative adversarial networks}
\newacronym{rdi}{RDI}{Range-Doppler image}
\newacronym{mira}{MIRA}{multi-class interpretable rule-based algorithm}
\newacronym{iqr}{IQR}{Interquatile range}
\newacronym{fft}{FFT}{fast Fourier transform}
\newacronym{rf}{RF}{radio-frequency}
\newacronym{if}{IF}{intermediate frequency}
\newacronym{std}{STD}{standard deviation}
\newacronym{rnn}{RNN}{recurrent neural network}
\newacronym{euai}{EU AI Act}{European Union's Artificial Intelligence Act}
\newacronym{lime}{LIME}{Local Interpretable Model-agnostic Explanations}
\newacronym{ai}{AI}{artificial intelligence}
\newacronym{vae}{VAE}{variational autoencoder}

\title{GenFacts-Generative Counterfactual Explanations for Multi-Variate Time Series}
\address{Author Affiliation(s)}
\name{\begin{tabular}{c}\textsuperscript{*}Sarah Seifi$^{1,2}$ \qquad \textsuperscript{*}Anass Ibrahimi$^{1,2}$ \qquad Tobias Sukianto$^{2,3}$ \qquad Cecilia Carbonelli$^{2}$ \\ 
\qquad Lorenzo Servadei$^{1}$  \qquad Robert Wille$^{1}$\end{tabular}}

\address{$^{1}$Technical University of Munich,
$^{2}$ Infineon Technologies AG, \\
$^{3}$Johannes Kepler University Linz}

\begin{document}
%

\maketitle

\begin{abstract}

Counterfactual explanations aim to enhance model transparency by showing how inputs can be minimally altered to change predictions. For multivariate time series, existing methods often generate counterfactuals that are invalid, implausible, or unintuitive. We introduce \textbf{GenFacts}, a generative framework based on a class-discriminative variational autoencoder. It integrates contrastive and classification-consistency objectives, prototype-based initialization, and realism-constrained optimization. We evaluate GenFacts on radar gesture data as an industrial use case and handwritten letter trajectories as an intuitive benchmark. Across both datasets, GenFacts outperforms state-of-the-art baselines in plausibility (+18.7\%) and achieves the highest interpretability scores in a human study. These results highlight that plausibility and user-centered interpretability, rather than sparsity alone, are key to actionable counterfactuals in time series data.




\end{abstract}


\section{Introduction}
\label{sec:introduction}
\renewcommand{\thefootnote}{\fnsymbol{footnote}} 
\footnotetext[1]{Equal contribution.}
\renewcommand{\thefootnote}{\arabic{footnote}} 

Deep learning classifiers for multivariate time series have achieved impressive performance across domains such as human activity recognition \cite{uddin2024deep}, healthcare \cite{morid2023time}, and radar-based gesture recognition \cite{seifi2025complying}. Despite their effectiveness, these models are often considered “black boxes”, limiting their adoption in sensitive and safety-critical domains where trust, accountability, and transparency are paramount \cite{seifi2025learning}. To mitigate this, the field of \ac{xai} has emerged, offering a diverse toolbox of methods to interpret and explain model predictions. Among these methods, counterfactual explanations are particularly compelling because they provide actionable feedback by answering: “What is the smallest change to the input that would have led to a different prediction?” \cite{molnar2024interpretable}. Unlike feature importance techniques such as SHAP \cite{lundberg2017unified} or LIME \cite{ribeiro2016should}, counterfactuals can suggest how users might adapt inputs to influence outcomes. This property makes them highly suitable for interactive systems, including gesture recognition and decision support.

However, generating user-friendly counterfactuals for multivariate time series remains a significant challenge. Early counterfactual methods, such as Wachter et al.’s optimization-based framework \cite{wachter2017counterfactual} and prototype-driven approaches \cite{van2021interpretable}, were developed for static tabular or image data and fail to capture temporal dependencies. More recent methods tailored to time series, e.g., CoMTE \cite{ates2021counterfactual}, CFProto \cite{van2021interpretable}, TSEvo \cite{hollig2022tsevo}, SPARCE \cite{lang2023generating}, and Multi-SpaCE \cite{refoyo2024multi},  improve on temporal consistency but still face three key limitations:
(i) Plausibility: Perturbation- or search-based approaches often generate samples that drift outside the realistic data manifold, violating physical or causal constraints of the domain; (ii) Feature correlations: Many methods encourage sparsity, but in multivariate time-series data (e.g., radar), features are inherently cross-time and cross-feature coupled (e.g., radial distance and velocity), making independent changes unrealistic; (iii) User actionability: Optimizing primarily for minimal changes can yield theoretically valid counterfactuals that are practically uninterpretable or impractical for end-users. As a result, existing methods often yield counterfactuals that are formally valid yet semantically meaningless. We argue that in realistic multivariate settings, sparse counterfactuals are not necessarily useful. Instead, what matters is whether the explanation is (i) plausible (faithful to the true data-generating distribution),  and (ii) actionable (providing insights that end-users can meaningfully act upon). This shift from minimality toward user-centric interpretability is crucial for deploying counterfactuals in interactive AI systems.

To address these challenges, we introduce GenFacts (\underline{\textbf{gen}}erative counter\underline{\textbf{fact}}ual\underline{\textbf{s}}), a generative, model-agnostic framework for producing realistic and interpretable counterfactuals for multivariate time series. Our contributions are:


\begin{itemize}
\item A \ac{vae}-based counterfactual generator that integrates prototype-based initialization and contrastive learning to improve convergence and class separation.
\item A multi-objective loss that explicitly balances validity, realism, and user-centric actionability, including a realism constraint to ensure counterfactuals remain within dense, physically coherent regions of the data manifold.
\item A human-centric interpretability study, introducing the Interpretability Score, where counterfactuals are rated based on whether they provide clear, actionable corrections to observed anomalies.
\item Comprehensive benchmarking against state-of-the-art approaches on two multivariate time series datasets, demonstrating superior performance in plausibility and interpretability.
\end{itemize}

Compared to state-of-the-art baselines, GenFacts improves plausibility by +18.7\%, maintains 100\% validity of generated counterfactuals, and achieves the highest interpretability scores in a human study.

The rest of this paper is structured as follows: Section~\ref{sec:methodology} details the proposed methodology. Section~\ref{sec:experiments} presents experimental results and comparisons. Section~\ref{sec:conclusion} concludes with a summary and future research directions.

\section{Methodology}
\label{sec:methodology}

GenFacts is a two-stage pipeline for generating plausible and actionable counterfactual explanations for multivariate time series classifiers (Fig.~\ref{fig:method}).  
Stage~1 trains a \gls{vae} to learn a class-relevant and structured latent space.  
Stage~2 performs a gradient-based search in this latent space to generate counterfactuals that meet user-defined criteria.

\begin{figure}[t]
\centering
\scalebox{0.72}{ 

\begin{tikzpicture}[
    font=\footnotesize, 
    node distance=4mm and 7mm, 
    >=Latex,
    data/.style={shape=rectangle,draw,minimum width=22mm,minimum height=5mm,fill=white,rounded corners=1pt,align=center},
    proc/.style={draw,rounded corners=2pt,minimum width=22mm,minimum height=7mm,fill=blue!6,align=center},
    frozen/.style={proc,fill=gray!15,dashed},
    loss/.style={circle,draw,minimum size=4mm,inner sep=0.3mm,fill=orange!12},
    group/.style={draw,rounded corners=3pt,inner sep=4pt},
    parallelogram/.style={
        shape=trapezium,
        trapezium left angle=60, 
        trapezium right angle=120, 
        draw,
        fill=white,
        inner xsep=2mm, 
        inner ysep=1mm, 
        align=center
    }
]

\begin{scope}[xshift=-60mm] 
\node (title1) {\textbf{Stage 1: VAE Training}};

\node (xin)   [parallelogram,below=of title1] {Input sample $x$};
 
\node (enc)   [proc,below=of xin] {Encoder\\(BiLSTM + 1D CNN + 1FC)};
\node (z)     [proc,below=of enc] {$z$ (latent vector)};
\node (Lkl)   [loss,left=3mm of z] {$L_{\mathrm{KL}}$};
\node (Lcon)  [loss,right=3mm of z] {$L_{\text{con}}$};
\node (dec)   [proc,below=of z] {Decoder\\(CNN$^{\top}$ + BiLSTM + 1FC)};
\node (xrec)  [parallelogram,below=of dec] {Reconstructed sample $x'$};
\node (Lrec)  [loss,left=3mm of xrec] {$L_{\text{rec}}$};
\node (clf)   [frozen,below=of xrec] {Frozen Classifier\\(GRU + FC)};
\node (ypred) [parallelogram,below=of clf] {Prediction};
\node (Lcls)  [loss,left=3mm of ypred] {$L_{\text{cc}}$};

\draw[->] (xin) -- (enc);
\draw[->] (enc) -- (z);
\draw[->,orange!80] (z) -- (Lcon);
\draw[->,orange!80] (z) -- (Lkl);
\draw[->] (z) -- (dec);
\draw[->] (dec) -- (xrec);
\draw[->,orange!80] (xrec) -- (Lrec);
\draw[->] (xrec) -- (clf);
\draw[->] (clf) -- (ypred);
\draw[->,orange!80] (ypred) -- (Lcls);
\end{scope}

\begin{scope}[xshift=0mm] 
\node (title2) [align=center] {\textbf{Stage 2: Counterfactual Generation}};
\node (xorig) [parallelogram,below=of title2] {Input sample $x$};
\node (enc2)  [frozen,below=of xorig] {Frozen Encoder\\(from Stage 1)};
\draw[->] (xorig) -- (enc2);


\node[draw=red!70!black, dashed, thick, rounded corners=4pt, fill=red!2,
      inner sep=8pt, minimum width=58mm, minimum height=17mm,
      below=of enc2] (optbox) {};
\node[text=red!80!black, anchor=north]
      at (optbox.north) {Gradient-Based Latent Optimization};

 \draw[->] (enc2) -- (optbox);     
\node (zinit) [proc, minimum width=15mm] at ([xshift=-18mm]optbox.center) {$z$ };
\node (zopt)  [proc, minimum width=15mm] at ([xshift= 18mm]optbox.center) {$z_{cf}$};
\draw[->] (zinit) -- (zopt) node[midway,below]{\scriptsize iterative updates};

\node (proto) [proc,left=3mm of enc2, label={[text=orange!85!black]}] {Target class prototype\\$\mu_{k\text{-NN}}$};
\draw[->] (proto) -- (optbox);

\node (Lreal) [loss,left=3mm of optbox] {$L_{\text{real}}$};
\draw[->,orange!80] (optbox) -- (Lreal);

\coordinate (redcenter) at (optbox.center);
\node (dec2)  [frozen,below=15mm of redcenter] {Frozen Decoder\\(from Stage 1)};
\draw[->] (optbox) -- (dec2);
\node (xcf)   [parallelogram,below=of dec2] {Counterfactual sample $x_{\text{cf}}$};
\draw[->] (dec2) -- (xcf);

\node (Lprox) [loss,right=3mm of xcf] {$L_{\text{prox}}$};
\draw[->,orange!80] (xcf) -- (Lprox);
\node (clf2)  [frozen,below=of xcf] {Frozen Classifier};
\draw[->] (xcf) -- (clf2);
\node (ycf)   [parallelogram,below=of clf2] {Prediction};
\draw[->] (clf2) -- (ycf);
\node (Lthr)  [loss,right=3mm of ycf] {$L_{\text{cls}}$};
\draw[->,orange!80] (ycf) -- (Lthr) node[midway,above=2mm]{};
\end{scope}


\node[group, left=20mm of ycf, minimum width=0.09\textwidth] (legend) {
\begin{tabular}{cc}
\tikz \node[proc,minimum width=5mm,minimum height=2.5mm]{}; learned module &
\tikz \node[frozen,minimum width=5mm,minimum height=2.5mm]{}; frozen module \\
\tikz \node[loss,minimum size=2.5mm]{}; loss term &
\tikz \node[parallelogram,rounded corners=0pt,minimum width=4mm,minimum height=2mm,inner xsep=1mm,inner ysep=0.5mm]{}; data
\end{tabular}
};

\end{tikzpicture}}
\caption{Overview of GenFacts, a model-agnostic two-stage framework. Stage~1 trains a VAE with multi-objective losses to learn a structured, class-discriminative latent space. Stage~2 generates counterfactuals via gradient-based latent optimization, initialized at a target-class prototype to improve plausibility and efficiency.}
\label{fig:method}
\end{figure}

\subsection{Stage 1: VAE Architecture and Training}

The core of our method is a VAE trained to learn a smooth, class-discriminative latent space that faithfully reconstructs time series data. The encoder pipeline consists of a 2-layer bidirectional LSTM followed by a 1D CNN stack to capture both long-range temporal dependencies and local features. A final fully connected layer projects these features into the latent distribution parameters, $\boldsymbol{\mu}$ and $\log \boldsymbol{\sigma}^2$. The decoder mirrors this architecture, using a fully connected layer and a series of transposed CNN layers to upsample the latent vector, followed by a bidirectional LSTM to reconstruct the time series $\hat{\mathbf{x}}$. We train the VAE with a multi-objective loss combining reconstruction fidelity, latent regularization, semantic consistency, and class separation:

\newlength{\desclen}
\setlength{\desclen}{0.38\linewidth} 

\begin{align}
\mathcal{L}_{\mathrm{rec}} &= \|\mathbf{x} - \hat{\mathbf{x}}\|_2^2, \; \text{\footnotesize \textit{\hspace{2.5cm}reconstruction fidelity}} \\[0.5mm]
\mathcal{L}_{\mathrm{KL}} &= D_{\mathrm{KL}}\!\big(\mathcal{N}(\boldsymbol{\mu},\boldsymbol{\sigma}^2)\,\|\,\mathcal{N}(0,\mathbf{I})\big), \; \text{\footnotesize \textit{\hspace{0.25cm}latent regularization}} \\[0.5mm]
\mathcal{L}_{\mathrm{cc}} &= H\!\big(P(\mathbf{y}|\mathbf{x}), P(\mathbf{y}|\hat{\mathbf{x}})\big), \; \text{\footnotesize \textit{\hspace{0.6cm}classification consistency}} \\[0.5mm]
\mathcal{L}_{\mathrm{con}} &= \mathrm{NT{\,}\text{-}{\,}Xent}(z_i, z_j). \; \text{\footnotesize \textit{\hspace{1.25cm}contrastive separation}}
\end{align}

Here, $\mathcal{L}_{\mathrm{rec}}$ ensures accurate reconstructions, $\mathcal{L}_{\mathrm{KL}}$ aligns the latent distribution with a standard normal prior, and $\mathcal{L}_{\mathrm{cc}}$ matches the frozen classifier’s softmax predictions on $\mathbf{x}$ and $\hat{\mathbf{x}}$. The contrastive loss $\mathcal{L}_{\mathrm{con}}$ uses the normalized temperature-scaled cross entropy (NT-Xent) to bring together latent vectors $z_i$ and $z_j$ from the same class while pushing apart those from different classes, improving class separability in the latent manifold \cite{chen2020simple}. The total VAE training objective is:

\begin{equation}
\mathcal{L}_{\mathrm{VAE}} = \lambda_{\mathrm{rec}}\mathcal{L}_{\mathrm{rec}} + \lambda_{\mathrm{KL}}\mathcal{L}_{\mathrm{KL}} + \lambda_{\mathrm{cc}}\mathcal{L}_{\mathrm{cc}} + \lambda_{\mathrm{con}}\mathcal{L}_{\mathrm{con}}.
\end{equation}

\subsection{Stage 2: Counterfactual Generation}
In the second stage, all modules from Stage 1 (the encoder, decoder, and classifier) are \emph{frozen} to ensure stability. Once the VAE is trained, we generate a counterfactual $\mathbf{x}_{\text{cf}}$ for a given input $\mathbf{x}_0$ and a target class $y_{\text{target}}$ by optimizing a latent vector $z_{\text{cf}}$ using gradient descent. Our optimization objective $\mathcal{L}_{\text{cf}}$ is designed to balance several key criteria:

\begin{align}
\mathcal{L}_{\mathrm{cls}} &= -\log P_{\theta}(y_{\mathrm{target}}\,|\,\mathbf{x}_{\mathrm{cf}}), \; 
\text{\footnotesize \textit{\hspace{1cm}target class match}} \\[0.5mm]
\mathcal{L}_{\mathrm{prox}} &= \|D(z_{\mathrm{cf}}) - \mathbf{x}_0\|_2^2, \; 
\text{\footnotesize \textit{\hspace{2.0cm}minimal change}} \\[0.5mm]
\mathcal{L}_{\mathrm{real}} &= \|z_{\mathrm{cf}} - \boldsymbol{\mu}_{\mathrm{prior}} \|_2^2, \; 
\text{\footnotesize \textit{\hspace{2cm}latent plausibility}}
\end{align}
where $D(\cdot)$ is the decoder and $\boldsymbol{\mu}_{\mathrm{prior}}$ is the zero-mean prior. The counterfactual objective is:
\begin{equation}
\mathcal{L}_{\mathrm{cf}} = \lambda_{\mathrm{cls}}\mathcal{L}_{\mathrm{cls}} + \lambda_{\mathrm{prox}}\mathcal{L}_{\mathrm{prox}} + \lambda_{\mathrm{real}}\mathcal{L}_{\mathrm{real}}.
\end{equation}

We initially optimized from the encoding of $\mathbf{x}_0$ with a prototype loss $\mathcal{L}_{\mathrm{proto}}$ to pull the counterfactual toward a target-class prototype. In practice, initializing $z_{\mathrm{cf}}$ directly at the prototype $\mu_{k\text{-NN}}$ (the mean latent vector of the $k$ nearest neighbors from the target class) yields higher success rates, faster convergence, and removes the need for $\mathcal{L}_{\mathrm{proto}}$ \cite{van2021interpretable}.

\subsection{Implementation Details and Training Configuration}

\textbf{VAE Configuration:} Our VAE was configured with a 2-layer bidirectional LSTM encoder and decoder, incorporating a Conv1D bottleneck. The latent dimension was set to 40. We used the Adam optimizer with ScheduleFree learning \cite{defazio2024road}, a learning rate of 0.002, and a batch size of 32. The model was trained for up to 600 epochs with optional early stopping, and validation was performed every 10 epochs.

\textbf{Loss Weights and Annealing:} The weights for VAE training were empirically determined based on validation loss as $\lambda_{\text{rec}}=1.0$, $\lambda_{\text{KL}}=0.01$, $\lambda_{\text{cc}}=1.0$, and $\lambda_{\text{con}}=0.5$. For counterfactual generation, we used $\lambda_{\text{cls}}=1.0$, $\lambda_{\text{prox}}=0.5$, and $\lambda_{\text{real}}=0.001$. The KL divergence was annealed via a cosine schedule, the classification and contrastive losses were linearly annealed over the first half of training.

\textbf{Plausibility Trade-offs:} Our method explicitly prioritizes plausibility over sparse changes. For multivariate time series like radar gestures, features are often physically interdependent. Enforcing sparse changes (e.g., via L1 loss) would often lead to implausible or unphysical counterfactuals. By operating in the VAE's latent space, our approach naturally generates explanations that respect these interdependencies, producing coherent and realistic explanations.

\section{Experimental Results}
\label{sec:experiments}

We validate our counterfactual generation framework through quantitative and qualitative analyses, ablation studies, and a human-centric evaluation. The goal is to assess whether our method produces counterfactuals that are not only valid and plausible, but also actionable for end-users.

\subsection{Datasets}
We evaluate on two multivariate time series datasets:  
(1) a real-world FMCW radar gesture dataset~\cite{radar-dataset}, containing five gestures (\textit{Push, Swipe Left/Right, Swipe Up/Down}), and  
(2) a handwritten letter trajectory dataset~\cite{character_trajectories_175}, where each sequence corresponds to a pen trajectory forming a character.  

For the radar dataset, we additionally recorded diagonal swipes (\textit{up-right, down-left}) to specifically test counterfactual interpretability. These ambiguous gestures lie between horizontal and vertical swipes and allow us to evaluate whether generated counterfactuals provide clear guidance to users.

Since the radar dataset reflects a more advanced, real-world setting with physically constrained sensor data, most of our detailed quantitative evaluation is performed on this dataset. The letter trajectory dataset, in contrast, is particularly well-suited for visualization and qualitative analysis, as its trajectories are directly interpretable as complete letters. We therefore use it primarily to visually illustrate how counterfactuals transform one sequence into another in a way that is human-understandable.

\subsection{Classifier Performance}
Our framework relies on well-performing classifiers.  
For radar gestures, we adopt the \ac{gru}-based sequence model and preprocessing pipeline from~\cite{strobel2024gesture}, extended with a moving average filter and normalization. The classifier was trained on 6,809 samples (80-20 train-validation split) using the Adam optimizer with ScheduleFree learning \cite{defazio2024road} for 80 epochs, achieving an F1 score of 0.98 on the validation set. 

For the handwritten letter trajectories (2,858 samples, 75-25 split), we trained a GRU classifier with almost identical optimization settings (8 hidden units, 100 epochs) using pen pressure and $(x,y)$ coordinates as input, achieving an F1 score of 0.92. These strong baseline classifiers provide a reliable foundation for counterfactual generation.

\subsection{Quantitative Evaluation}
We assess counterfactual quality using three standard metrics:  
\textbf{Validity} (percentage of counterfactuals classified as the target class),  
\textbf{Proximity} (L1 distance between original and counterfactual), and \textbf{Plausibility} (a counterfactual is plausible if it is (i) classified as an inlier by an ensemble of isolation forests and also (ii) lies within the 90$^{\text{th}}$ percentile of pairwise training distances) \cite{delaney2021instance,dantas2023counterfactual,filali2022mining}.  
For proximity and plausibility, samples are transformed into fixed-length descriptors (statistical moments, temporal dynamics, correlations) to ensure comparability. Table~\ref{tab:plausibility-results} reports results on the gesture radar dataset. GenFacts achieves the highest plausibility (\textbf{87.5\%}), significantly outperforming COMTE (68.8\%), while maintaining perfect validity. CFProto produces the most minimal changes (lowest proximity) but with extremely low plausibility and validity, making its counterfactuals unrealistic. This highlights the trade-off: methods prioritizing minimality often generate impractical edits, whereas GenFacts, by constraining counterfactuals to a structured latent space, produces explanations that are both realistic and actionable.  

\begin{table}[h!]
\centering
\resizebox{\columnwidth}{!}{%
\begin{tabular}{|l|c|c|c|}
\hline
\textbf{Method} & \textbf{Proximity $\downarrow$} & \textbf{Validity (\%)$\uparrow$} & \textbf{Plausibility (\%)$\uparrow$} \\
\hline
CoMTE     & 63.4   & 100.0 & 68.8 \\
TSEvo     & 50.6   & 100.0 & 3.1  \\
CFProto   & \textbf{1.8} & 52.5  & 0.0  \\
SPARCE    & 450.6  & 94.5   & 0.0  \\
Multi-SpaCE& 28.8  & 100.0 & 71.9 \\
\textbf{GenFacts} & 97.5  & \textbf{100.0} & \textbf{87.5} \\
\hline
\end{tabular}}
\caption{Quantitative results on the FMCW radar dataset. GenFacts achieves the best plausibility and perfect validity, highlighting the importance of plausibility over minimal edits.}
\label{tab:plausibility-results}
\end{table}

\subsection{Human-Centric Interpretability}

To assess whether counterfactuals are actionable for end-users, we introduce the Interpretability Score. Five independent evaluators labeled counterfactuals as "interpretable" if they clearly corrected the anomaly in the original gesture and indicated how the motion should be adjusted toward the target class. We focused on deliberately ambiguous diagonal swipes in the radar dataset. For example, when a user performed an \textit{Up-Right Swipe} instead of the intended \textit{Swipe Right}, the counterfactual should illustrate, e.g., via 3D visual overlays, how the motion must be adjusted horizontally to yield the correct prediction.  

As shown in Table~\ref{tab:interpretability-results}, GenFacts achieves the highest Interpretability Score (\textbf{90.4\%}), whereas COMTE reaches only 38.6\% and CFProto, TSEvo, and others fail almost completely. This large discrepancy can be explained by the fact that GenFacts explicitly optimizes for plausibility and realism in the latent space, producing counterfactuals that remain smooth and physically consistent with the data manifold. In contrast, copy–paste or sparsity-driven approaches often yield fragmented or unrealistic trajectories that evaluators cannot easily interpret. These results demonstrate that traditional metrics such as proximity do not necessarily translate into actionable, user-understandable explanations.  

\begin{table}[h!]
\centering
\begin{tabular}{|l|c|}
\hline
\textbf{Method} & \textbf{Interpretability Score (\%)} \\
\hline
CoMTE     & 38.6 \\
TSEvo     & 3.4 \\
CFProto   & 2.1 \\
SPARCE    & 9.7   \\
Multi-SpaCE   & 5.5 \\
GenFacts  & \textbf{90.4} \\
\hline
\end{tabular}
\caption{Interpretability Score across methods on diagonal radar gestures. GenFacts clearly outperforms baselines, highlighting that low-distance counterfactuals are not necessarily interpretable or actionable.}
\label{tab:interpretability-results}
\end{table}

\subsection{Qualitative Results}

\begin{figure}[h!]
    \centering
    \includegraphics[width=\columnwidth]{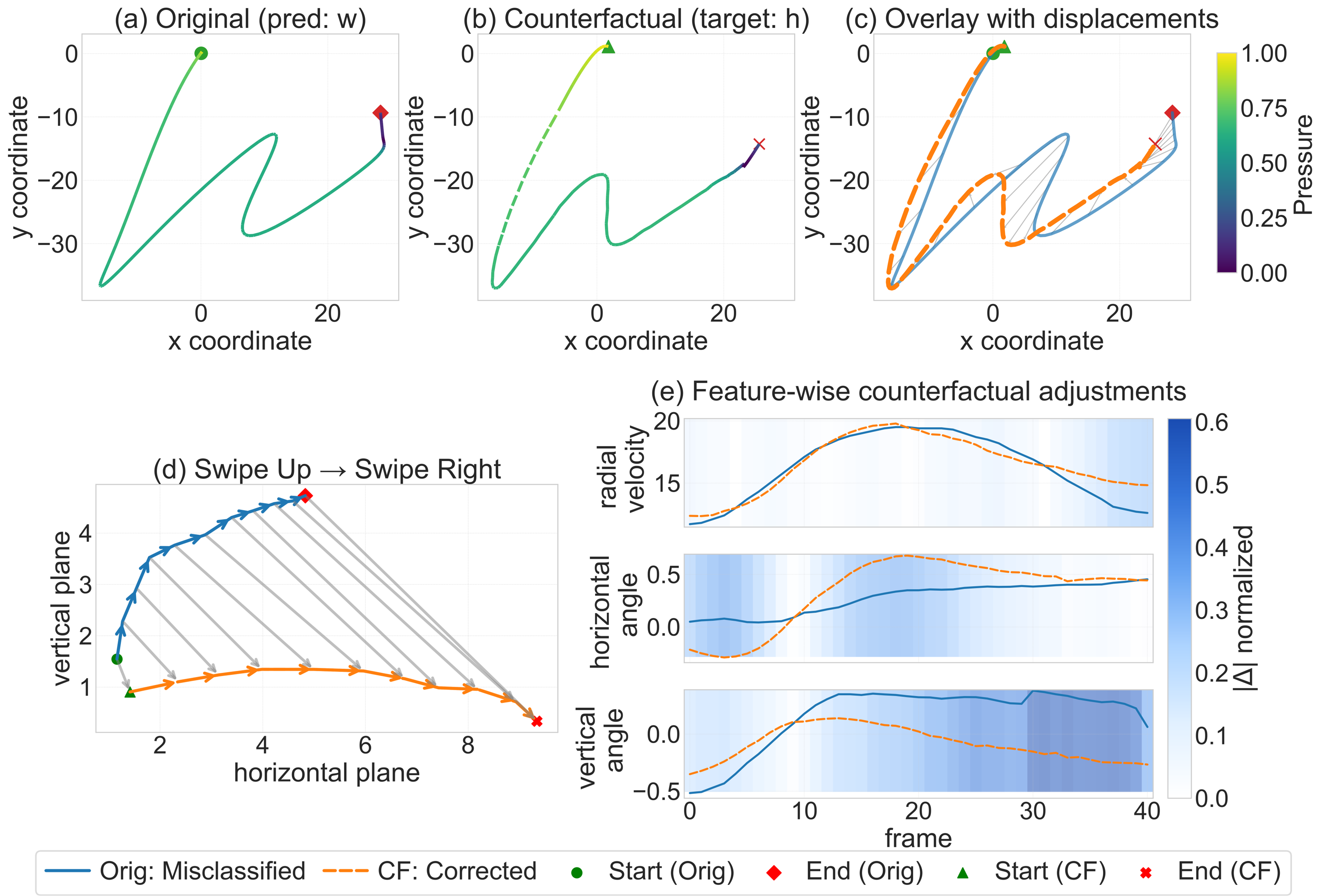}
    \caption{Qualitative results of GenFacts. (a–c) Letter trajectory: Counterfactual corrects a misclassified “\textit{h}” (predicted as “\textit{w}”). (d–e) Radar gesture: A diagonally performed \textit{Swipe Right}, misclassified as \textit{Swipe Up}, is corrected by flattening the vertical angle and increasing horizontal curvature.}
    \label{fig:results-figure}
\end{figure}

As shown in Fig.~\ref{fig:results-figure}, GenFacts yields coherent, semantically meaningful counterfactuals: smooth letter corrections (panels a–c) and intuitive radar gesture adjustments (panels d–e).

\subsection{Ablation Study}

We assess the role of each loss term by removing it in turn (Table~\ref{tab:ablation-results}).  
Without the proximity loss $\mathcal{L}_{\text{prox}}$, counterfactuals drift further from the input (proximity +96.7).  
Dropping the classification loss $\mathcal{L}_{\text{cls}}$ reduces validity by –18.7pp, while removing the realism loss $\mathcal{L}_{\text{real}}$ decreases plausibility by –18.7pp.  
These results confirm that each objective contributes complementary benefits: proximity for minimality, classification for target alignment, and realism for staying on the data manifold.

\begin{table}[h!]
\centering
\scriptsize   
\begin{minipage}{0.3\linewidth}
\centering
\begin{tabular}{lc}
\toprule
\multicolumn{2}{c}{Proximity $\downarrow$} \\
\midrule
Full & 97.5 \\
w/o $\mathcal{L}_{\mathrm{prox}}$ & 194.2 \\
$\Delta$ (worse) & $+96.7$ \\
\bottomrule
\end{tabular}
\end{minipage}
\hspace{0.9em}
\begin{minipage}{0.3\linewidth}
\centering
\begin{tabular}{lc}
\toprule
\multicolumn{2}{c}{Validity (\%) $\uparrow$} \\
\midrule
Full & 100.0 \\
w/o $\mathcal{L}_{\mathrm{cls}}$ & 81.3 \\
$\Delta$ (pp) & $-18.7$ \\
\bottomrule
\end{tabular}
\end{minipage}
\hspace{0.9em}
\begin{minipage}{0.3\linewidth}
\centering
\begin{tabular}{lc}
\toprule
\multicolumn{2}{c}{Plausibility (\%) $\uparrow$} \\
\midrule
Full & 87.5 \\
w/o $\mathcal{L}_{\mathrm{real}}$ & 68.8 \\
$\Delta$ (pp) & $-18.7$ \\
\bottomrule
\end{tabular}
\end{minipage}
\caption{Ablation results for GenFacts. Each block reports the full model, the relevant ablation, and the deterioration $\Delta$ (positive means worse for proximity; negative pp means a drop for percentages).}
\label{tab:ablation-results}
\end{table}

\section{Conclusion}
\label{sec:conclusion}

We presented GenFacts, a generative framework for producing realistic and interpretable counterfactuals in multivariate time series. By combining a class-discriminative VAE, prototype-based initialization, and contrastive learning within a multi-objective loss, GenFacts achieves a +18.7\% improvement in plausibility, maintains 100\% validity, and obtains the highest interpretability scores in a human study. Evaluations on radar gesture and letter trajectory datasets confirm that proximity-based metrics alone are insufficient, underscoring the importance of plausibility and human-centric evaluation for trustworthy explanations. Future work could extend GenFacts to streaming data, multimodal sensing, and causality-aware counterfactuals.



\bibliographystyle{IEEEbib}
\bibliography{strings,refs}

\end{document}